\documentclass{article}
\usepackage{amsmath,amssymb,amsfonts}

\usepackage{natbib}
\usepackage{booktabs}
\setcitestyle{authoryear,open={(},close={)}}

\usepackage{rotating}
\usepackage{fancyvrb}
\usepackage[utf8]{inputenc} % allow utf-8 input
\usepackage[T1]{fontenc}    % use 8-bit T1 fonts
\usepackage{url}            % simple URL typesetting
\usepackage{nicefrac}       % compact symbols for 1/2, etc.
\usepackage{microtype}      % microtypography
\usepackage{xcolor}         % colors
\usepackage{algorithm}
\usepackage{algpseudocode}
\usepackage{graphicx}
\usepackage{dsfont}
\usepackage{pdfpages}

\usepackage{multirow}%
\usepackage{amsthm}%
\usepackage{mathrsfs}%
\usepackage[title]{appendix}%
\usepackage{textcomp}%
\usepackage{manyfoot}%
\usepackage{booktabs}%
\usepackage{algorithmicx}%
\usepackage{listings}%
\usepackage{caption}
\usepackage{subcaption}

\usepackage{csquotes}
\usepackage{makecell}

\usepackage{hyperref}       % hyperlinks
\usepackage{bookmark}

\title{Discovering the curriculum with AI:\\
A proof-of-concept demonstration\\
with an intelligent tutoring system\\
for teaching project selection}

\author{Lovis Heindrich$^1$$^*$, Falk Lieder$^{1, 2}$}
\date{
    $^1$Max Planck Institute for Intelligent Systems, Tübingen, Germany\\
    $^2$Department of Psychology, UCLA, Los Angeles, USA\\
    $^*$E-mail: lovis.heindrich@tuebingen.mpg.de\\[2ex]
    \today}

\begin{document}

\maketitle

\abstract{% Now 172 words, max is 200 words
The decisions of individuals and organizations are often suboptimal because fully rational decision-making is too demanding in the real world. Recent work suggests that some errors can be prevented by leveraging artificial intelligence to discover and teach clever heuristics. So far, this line of research has been limited to simplified, artificial decision-making tasks. This article is the first to extend this approach to a real-world decision problem, namely, executives deciding which project their organization should launch next. We develop a computational method (MGPS) that automatically discovers project selection strategies that are optimized for real people, and we develop an intelligent tutor that teaches the discovered project selection procedures. We evaluated MGPS on a computational benchmark and tested the intelligent tutor in a training experiment with two control conditions. MGPS outperformed a state-of-the-art method and was more computationally efficient. Moreover, people who practiced with our intelligent tutor learned significantly better project selection strategies than the control groups. These findings suggest that AI could be used to automate the process of discovering and formalizing the cognitive strategies taught by intelligent tutoring systems.
}

\maketitle
\newpage

\section{Introduction}

Teaching general cognitive skills is an essential goal of education \citep{collins1991cognitive}. One way to represent cognitive skills is through cognitive strategies. For example, when teaching how to effectively solve math problems, instructors might teach their students metacognitive strategties such as evaluating the effectiveness of different solution methods or breaking problems into smaller subproblems. Prior work has shown that teaching students skills such as carefully evaluating their problem understanding of math problems can significantly improve their test scores \citep{ozsoy2009effect}. Similarly, students' writing quality can be improved by teaching them strategies for planning and revising their texts \citep{graham2007meta}. 

The teaching of cognitive strategies can be automated by  developing intelligent tutoring systems \citep{anderson1985intelligent,koedinger1997intelligent,graesser2012intelligent}. 
In current applications, the strategies the intelligent tutoring systems teach have to be specified manually \citep{aleven2009new, chi2010meta, ozsoy2009effect}. While hand-crafted strategies can often be effective, they rely on the designer's intuitions. This constitutes a key bottleneck, as it presumes that designers know the optimal cognitive strategies and can precisely articulate them. While this may be the case in some limited domains, in many domains it is unknown which cognitive strategies are optimal. Expertise in areas such as mathematical problem solving often involves tacit, procedural knowledge that is difficult to verbalize \citep{anderson1982acquisition}.  
One way to overcome this bottleneck is to develop rational process models of human cognition \citep{LiederGriffiths2020}. Recent technical advances have made it possible to automate this process by leveraging AI to discover near-optimal cognitive strategies automatically \citep{Callaway2021Leveraging,krueger2024identifying}. Adopting the strategies discovered by these methods consistently allowed people to perform better than the strategies they intuitively used \citep{Callaway2021Leveraging, consul2022, heindrich2025intelligent}.
Moreover, it is now possible to develop intelligent tutotring systems that use AI to discover the cognitive strategies they teach by themselves \citep{Callaway2021Leveraging}. However, so far, this approach has only been evaluated in artificial decision-making tasks. Therefore, its viability for educational applications is still unclear.

In this article, we present the first proof-of-concept that automatic strategy discovery can inform educational applications in the real world.
%a real educational example by demonstrating how a specific cognitive problem can be modeled from real-world data, how it is possible to algorithmically discover efficient cognitive strategies, and how intelligent tutoring systems can teach these strategies to people in an automated way. This contrasts with previous approaches that rely on hand-designed cognitive strategies through the utilization of automated strategy discovery methods. 
We build directly on our prior work \citep{heindrich2025intelligent,consul2022}, which brought strategy discovery closer to real-world application by introducing a metareasoning model of planning in partially observable environments. Although the planning task was complex, it was still highly artificial and unrealistic. Here,  we show that the general approach to combine automatic strategy discovery with an intelligent tutor can be extended to an important class of real-world decisions: deciding which project an organization, team, or individual should pursue next. This class of decision problems is known as \textit{project selection}. 

%can be extended this approach by modeling decision-making in the real-world in the same mathematical framework, using strategy discovery methods to discover efficient cognitive strategies, and teaching people the discovered strategies using an intelligent tutor. 

%The cognitive skill to which we apply this method is project selection. 
Corporations and individuals often need to choose a project from multiple alternatives. These project selection problems are usually high-stakes decisions that can be highly impactful for the future of an organization. For example, an organization looking for a sustainable investment project \citep{Khalili-Damghani_Sadi-Nezhad_2013} can benefit both financially and by improving its public image by selecting an impactful and profitable project, or incur major losses by selecting an unsuitable project.

Previous research on project selection recommends that candidate projects should be evaluated by multiple experts \citep{Coldrick_Lawson_Ivey_Lockwood_2002, Khalili-Damghani_Sadi-Nezhad_2013, liu_evaluation_2017}, and many structured approaches to integrate the experts' opinions exist \citep{de_souza_mcdm-based_2021}. However, existing project selection techniques are not well utilized in the real world \citep{schmidt_recent_1992, liu_evaluation_2017, Henriksen_Traynor_1999}. We hypothesize this underutilization is partly caused by the complexity and implementation costs of normative project selection strategies. Scoring approaches, which score alternative projects on several relevant selection criteria, have been proposed as a less complex selection method that can be applied to real-world problems more easily \citep{Henriksen_Traynor_1999, Khalili-Damghani_Sadi-Nezhad_2013, Coldrick_Lawson_Ivey_Lockwood_2002, Sadi-Nezhad_2017}. However, even scoring approaches face significant challenges when applied to real-world problems: precisely evaluating projects on multiple criteria can be a time-consuming and expensive process, but existing selection techniques often pay little attention to the information-gathering costs and instead assume expert opinions on all criteria are readily available and instead focus on how to integrate existing expert evaluations into a final project choice \citep{abdel2019hybrid, Khalili-Damghani_Sadi-Nezhad_2013}.

Due to the limited utilization of complex project selection methods, decision-makers, therefore, often rely on their intuition and much simpler techniques like brainstorming \citep{kornfeld_selection_2013}. 
This is concerning because the intuitive decisions of groups and individuals are highly susceptible to biases and unsystematic errors \citep{kahneman1982judgment}, such as weighing one's own opinion too strongly \citep{Yaniv_Kleinberger_2000}, being overconfident, and weighting the recommendations of advisors by their confidence rather than their competence \citep{Bonaccio_Dalal_2006}.
%When deciding with the help of multiple advisors, as is often the case for project selection problems, these biases can cause the decision-maker to choose badly. 
%Common biases in group decision-making include weighing one's own opinion too strongly \citep{Yaniv_Kleinberger_2000}, being overconfident, and weighting the recommendations of advisors by their confidence rather than their competence \citep{Bonaccio_Dalal_2006}.
However, people's errors in decision-making can partly be prevented by teaching prescriptive decision strategies adapted to humans by taking their cognitive limitations into account \citep{LiederGriffiths2020, Callaway2021Leveraging}. This approach to improving human decision-making is known as \textit{boosting} \citep{hertwig2017nudging}. 

To identify appropriate decision strategies, we can score candidate strategies by their \textit{resource-rationality}, that is, the degree to which they make rational use of people's limited time and bounded cognitive resources  \citep{LiederGriffiths2020}. In the resource-rational framework, the decision operations people can perform to arrive at a decision are modeled explicitly and assigned a cost. The overall efficiency of a decision strategy $h$ in an environment $e$ can then be computed by subtracting the expected costs $\lambda$ of the $N$ used decision operations from the expected utility $R_{total}$ of the resulting decision (see Equation~\ref{eq:rr}) \citep{consul2022}. This measure is called resource-rationality score (\textit{RR-score}) \citep{consul2022}. %It will be used throughout this article to evaluate and compare alternative decision strategies. 
People are usually not fully resource-rational, and identifying decision strategies would enable people to perform as well as possible is an important open problem \citep{Callaway2021Leveraging, consul2022, heindrich2025intelligent, mehta2022leveraging}.

\begin{equation}
\text{RR}(h,e)=\mathbb{E}[R_{total}|h, e] - \lambda \mathbb{E}[N | h, e]
\label{eq:rr}
\end{equation}

Recent work has demonstrated that the theory of resource rationality makes it possible to leverage AI to automatically discover and teach decision strategies that enable real people to make their decisions as well as possible \citep{Callaway2021Leveraging,consul2022,becker2022boosting,skirzynski2021automatic,mehta2022leveraging}. This approach has been dubbed \textit{AI-powered boosting}. %derives resource rationality as a theoretical foundation for boosting people's decision-making competence.
The first step of AI-powered boosting is to compute resource-rational decision strategies. Automatic strategy discovery methods \citep{Callaway2021Leveraging, consul2022, heindrich2025intelligent, skirzynski2021automatic,mehta2022leveraging} can discover efficient decision strategies by solving the metareasoning problem of deciding which decision operations to perform. While recent work has extended automatic strategy discovery methods to larger \citep{consul2022} and partially observable environments \citep{heindrich2025intelligent}, so far, they have not been applied to real-world decisions such as project selection. 

%Developing computational methods for decision support is one of the core problems of operations research \citep{gupta2022artificial, eom2006survey}.
In this article, we present a novel decision support method with the goal of discovering prescriptive decision-strategies that can be taught to people to improve how they select projects. Project selection is challenging because many crucial attributes of the candidate projects, such as their expected profitability, cannot be observed directly. Instead, they have to be inferred from observations that are not fully reliable. We, therefore, formalize project selection strategies as policies for solving a particular class of partially observable Markov Decision Processes (POMDPs). This formulation allows us to develop the first algorithm for discovering resource-rational strategies for human project selection. % by extending the only existing algorithm for discovering resource-rational decision strategies for partially observable environments\citep{heindrich2025intelligent} to the project selection problem.
To achieve this, we model a realistic project selection task as a metareasoning problem. The metareasoning consists in deciding which information one should request from which advisors when information is highly limited, uncertain, and costly. We develop an efficient algorithm for solving this problem based on MGPO, an algorithm for discovering resource-rational strategies for human decision-making in partially observable environments \citep{heindrich2025intelligent}, and apply it to derive a prescriptive decision strategy for a project selection problem a financial institution faced in the real world \citep{Khalili-Damghani_Sadi-Nezhad_2013}. %by extending the best existing algorithm for discovering resource-rational strategies for human decision-making in partially observable environments \citep{heindrich2025intelligent}. 
%We further develop a novel strategy discovery method based on MGPO \citep{heindrich2025intelligent} that can automatically discover near-optimal decision strategies for the project selection problem. 
Finally, we develop an intelligent tutor \citep{Callaway2021Leveraging} that teaches the decision strategy discovered by our method to people. We evaluated our approach by letting our intelligent tutor teach the automatically discovered project selection strategy to about 100 people, and then evaluated the quality of their decisions in realistic project selection problems against two control groups. Our results indicate that our approach can successfully improve human decisions in real-world problems where people are reluctant to let machines decide for them. 

\section{Background}

% We probably won't need separate sections for every topic here

%\subsection{Project selection}
\paragraph{Project selection}

In the project selection problem, a decision-maker aims to select the best-fitting project out of several candidates \citep{Sadi-Nezhad_2017}. Apart from a project's profitability, the evaluation usually also considers other factors, such as the alignment with organizational goals  \citep{carazo_project_2012}. This problem can be formalized as multi-criteria decision-making (MCDM) \citep{de_souza_mcdm-based_2021, mohagheghi_project_2019}. Projects can be evaluated using a scoring technique, which evaluates relevant criteria and then combines them to a weighted sum \citep{Sadi-Nezhad_2017}. Common approaches to solving the project selection problem include techniques such as the analytic hierarchy process, the analytic network process, real options analysis, and TOPSIS (see \citep{de_souza_mcdm-based_2021} for a review). These methods are commonly combined with fuzzy sets to account for uncertainty \citep{Khalili-Damghani_Sadi-Nezhad_2013}. However, these methods are rarely used in real-world problems because implementing them would require gathering and integrating a lot of information through a time-consuming process, which is often incompatible with the organizational decision process \citep{schmidt_recent_1992, liu_evaluation_2017}. Instead, organizations often rely on simpler, less structured methods like brainstorming \citep{kornfeld_selection_2013}. In addition, while the detailed information required by these methods can be costly and time-consuming to acquire in real-world settings, the methods often don't take into account how much information is needed to make an efficient decision.

\paragraph{Judge-advisor systems}
%We frame the project selection problem as a Judge-Advisor System (JAS) \citep{Bonaccio_Dalal_2006}.
In a Judge Advisor System (JAS) \citep{Bonaccio_Dalal_2006}, typically, a single decision-maker has to make a decision, and multiple advisors support the decision-maker by offering advice. Variations of the task can include costly advice \citep{Yaniv_Kleinberger_2000, gino2008we}, or advisors with varying reliability \citep{Olsen_Roepstorff_Bang_2019}. This is a common situation when CEOs decide which project their company should launch next. Unfortunately, decision-makers are known to be highly susceptible to systematic errors, such as weighing one's own opinion too strongly, overconfidence, egocentric advice discounting, and weighting the recommendations of advisors by their confidence rather than their competence \citep{Bonaccio_Dalal_2006, Ronayne_Sgroi_others_2019, Yaniv_Kleinberger_2000}. We model the project selection problem within the JAS framework by letting project evaluators take the role of advisors with varying reliability.

\paragraph{Resource rationality}

Resource-rational analysis is a cognitive modeling paradigm used to describe optimal decision-making with limited computational resources \citep{LiederGriffiths2020}. The paradigm considers not only the quality of decisions, but also the computational costs of reaching them. According to this paradigm, good decision-making consists in making efficient use of the available computational resources by balancing the competing objectives of maximizing the expected decision quality and minimizing computational costs. Resource rationality can be used to describe the quality of a decision strategy by subtracting the cost of the cognitive operations from the expected reward of the resulting decision. This measure is called resource-rationality score (\textit{RR-score}) \citep{consul2022}. It will be used throughout this article to evaluate and compare alternative decision strategies. People are usually not fully resource-rational, and figuring out what decision strategies would enable people to perform as well as possible is an important open problem \citep{Callaway2021Leveraging, consul2022, heindrich2025intelligent, mehta2022leveraging}.

\paragraph{Strategy discovery methods}

Discovering resource-rational planning strategies can be achieved by solving a meta-level Markov decision process \citep{hay2014selecting, callaway2017learning, callaway2018resource}, which models the metareasoning process as a Markov Decision Process (MDP), which state represents the current belief about the environment and actions represent decision operations. Performing a decision operation results in a negative cost and results in an update to the belief state. A special termination action represents exiting the metareasoning process and making a real-world decision, guided by the current beliefs \citep{hay2014selecting}. Multiple methods for solving meta-level MDPs exist (e.g. \citep{Russell_Wefald_1991, callaway2017learning, hay2014selecting, consul2022}). We refer to these algorithms as strategy discovery methods \citep{callaway2017learning,Callaway2021Leveraging,skirzynski2021automatic,consul2022,heindrich2025intelligent,mehta2022leveraging}. They learn or compute policies for selecting sequences of cognitive operations (i.e., computations) people can perform to reach good decisions. 

%In this article, we propose an adapted version of 
MGPO \citep{heindrich2025intelligent} is currently the only strategy discovery algorithm that can efficiently approximate resource-rational strategies for decision-making in partially observable environments. MGPO chooses decision operations by approximating their value of computation \citep{Russell_Wefald_1991} in a myopic manner: it always selects the computation whose execution would yield the highest expected gain in reward if a decision had to be made immediately afterward, without any additional planning.
MGPO makes use of the myopic approximation due to the high computational complexity of partially observable Markov Decision Processes (POMDP). Alternative approaches to modeling information gathering POMDPs are difficult to apply to partially observable meta-level MDPs, as they require carefully designed information rewards \citep{boutilier2002pomdp} or simplify the problem by restricting environments to be symmetric \citep{doshi2008permutable}. 

\paragraph{Intelligent tutoring systems}

Intelligent tutoring systems (ITS) are automated computer systems that teach skills in digital learning environments \citep{graesser2012intelligent}. They provide a highly scalable learning environment \citep{koedinger1997intelligent} that often can be similarly effective or even more effective than human tutors \citep{vanlehn2011relative}. ITS achieve this by offering a personalized learning environment in which learners are able to practice skills interactively, often supported through a cognitive model of the learner's progress that allows to adapt the learning process to their current needs \citep{woolf2010building}. Recently, ITS have also been applied to teach general cognitive skills such as help-seeking \citep{aleven2006toward} or reading comprehension \citep{guerra2017book}. 

\paragraph{Cognitive tutors}

Past work has developed cognitive tutors that teach automatically discovered planning strategies to people \citep{Callaway2021Leveraging,skirzynski2021automatic,mehta2022leveraging,consul2022}. Training experiments indicated that training with these cognitive tutors could significantly boost the quality of people's planning and decision-making  \citep{ Callaway2021Leveraging, skirzynski2021automatic, consul2022, heindrich2025intelligent,mehta2022leveraging}. These cognitive tutors teach efficient decision strategies in an automated manner, usually by computing the value of available decision operations using strategy discovery methods, and providing the learner feedback on the quality of the computations they select. 
%Initial work on improving human decision-making was limited to small planning tasks due to the computational complexity of solving the corresponding meta-level MDPs \citep{CognitiveTutorsRLDM, Callaway2021Leveraging}. While recent work has extended existing methods to much larger \citep{consul2022} and partially observable \citep{heindrich2025intelligent} settings, none of these methods have been applied to naturalistic problems so far.
Initially limited to small planning tasks due to the computational complexity of solving meta-level MDPs \citep{CognitiveTutorsRLDM, Callaway2021Leveraging}, recent work has extended existing methods to larger \citep{consul2022} and partially observable \citep{heindrich2025intelligent} settings. However, none of these methods have been applied to naturalistic problems so far.

A crucial obstacle is that these methods are limited to settings where 
all decision-relevant information comes from the same source. By contrast, in the real world, people have to choose between and integrate multiple different sources of information. In doing so, they have to take into account that some information sources are more reliable than others. Additionally, current strategy discovery methods are limited to artificial settings where each piece of information is an estimate of a potential future reward. By contrast, in the real world, most information is only indirectly related to future rewards, and different pieces of information have different units (e.g., temperature vs. travel time).  
%all information to be given in the same unit of measurement so that alternative aspects can be weighed against each other directly.
%Both of these assumptions are likely to be broken in real-world scenarios, where information can come from wildly different sources with different reliabilities.

%\section{Method}

\section{Formalizing optimal decision strategies for human project selection as the solution to a meta-level MDP}

In this section, we introduce our general resource-rational model of project selection, which we expect to be widely applicable to concrete, real-world project selection problems.
%, and then introduce the specific project selection task we implemented after \citet{Khalili-Damghani_Sadi-Nezhad_2013}, who work in a concrete real-world project selection setting where a decision-maker has to select between five project alternatives. 

Our model of project selection consists of two decision problems, an object-level decision-problem and a meta-level MDP \citep{callaway2017learning, hay2014selecting}. The two decision problems separate the actions the decision-maker has to choose between (object level), such as executing one project versus another, from decision operations that represent thinking about which of those object-level actions to perform (meta-level), such as gathering information about the projects' attributes. This allows us to solve both problems separately. 
%Formally, the object-level MDP consists of a set of $N_P$ available projects $\mathcal{P}=\{p_1,...p_{N_P}\}$. 
The object-level decision problem is a MCDM problem, where a set of $N_P$ potential projects $\mathcal{P}=\{p_1,...p_{N_P}\}$ are evaluated using $N_C$ relevant criteria $C=[c_1, ...c_{N_C}]$ weighted by fixed predetermined weights $W=[w_1, ...w_{N_C}]$. Actions in the object-level problem represent selecting the corresponding project ($\mathcal{A}=\{a_1,...,a_{N_P}\}$). The reward of selecting a project is computed by summing the weighted criteria scores of the selected project: $r^O(a_i)=\sum_{c}w_cc_{c, i}$ \citep{Coldrick_Lawson_Ivey_Lockwood_2002}. 

While the object-level decision is to select a project, the meta-level MDP is our formalization of the problem of discovering resource-rational project selection strategies. A project selection strategy is a systematic procedure for gathering decision-relevant information about various projects and turning it into a decision. We model this information gathering as a sequence of decisions about whether to request additional information and, if so, which information to request next. In each step, the decision-maker can ask one expert to rate one feature of one project. They can then use the resulting information to decide which, if any, additional information to request in the next step. Different project selection strategies differ in 1) which information they request depending on what is already known, and 2) when they stop gathering information and select a project based on what is already known. Resource-rational project selection strategies make these choices near optimally to achieve the best possible trade-off between the quality of the selected project and the cost of the time and information invested to select it. 

States in the meta-level MDP are belief states that represent the current information about each project's attributes. We model belief states using a multivariate Normal distribution to quantify the estimated value and uncertainty about the $N_P$ projects' scores on the $N_C$ criteria: $b=\left[ \left(\mu_{1,1}, \sigma_{1,1}\right), \cdots, \left(\mu_{N_P,N_C}, \sigma_{N_P,N_C}\right)\right]$. The actions (decision operations) of the meta-level MDP gather information about the different attributes of projects by asking one of the $N_E$ experts for their estimate of how one of the projects scores on one of the criteria. Experts provide discrete estimates from $min_{obs}$ to $max_{obs}$, and expert estimates can differ in their reliability and their cost. Specifically, the available actions are $A^M=\left\{a_{1, 1, 1}, \cdots, a_{N_P, N_C, N_E}, \perp\right\}$, where the meta-level action $a_{i, j, k}$ represents asking the expert $e_k$ for their estimate of criterion $c_j$ of project $p_i$. 
%updated via Bayesian inference, taking the reliability of the expert $\sigma_e$ into account 
%see Equations~\ref{eq:belief_update_mu} and~\ref{eq:belief_update_sigma}). 

\begin{align}
\hat{\mu} &\gets \left(\frac{w_{c_i} \cdot \mu}{(w_{c_i} \cdot \sigma)^2}+\frac{w_{c_i} \cdot obs}{(w_{c_i} \cdot \sigma_e)^2} \right) \cdot \left((w_{c_i} \cdot \sigma)^2+(w_{c_i} \cdot \sigma_e)^2 \right)
\label{eq:belief_update_mu} \\
\hat{\sigma} &\gets \sqrt{\frac{1}{\frac{1}{(\sigma\cdot s)^2}+\frac{1}{(\sigma_e \cdot s)^2}}}
\label{eq:belief_update_sigma}
\end{align}

Equations~\ref{eq:belief_update_mu} and~\ref{eq:belief_update_sigma} describe the belief update. After receiving information $obs$ from an expert, the current belief state $b_{p_ic_j}=\mathcal{N}(\mu, \sigma)$ is updated by applying the conjugate Gaussian update, which integrates the new observation $obs$ with standard deviation $\sigma_e$ (the expert's reliability) into the current belief. Beliefs are scaled by their respective criteria weight $w_c$ throughout. 

The reward of these meta-level actions is the negative cost $r^M(a_{i, j, k})=-\lambda_{e_k}$ of asking the expert $e_k$ for help. Finally, the meta-level action $\perp$ is the termination action, which corresponds to terminating the decision-making process and selecting a project. The reward of the termination action is the expected reward of selecting the best project according to the current belief state. An optional budget parameter $N_a$ specifies the maximum number of meta-level actions that can be performed before the termination operation is selected. 

Meta-level MDPs are notoriously difficult to solve due to their extremely large state space \citep{consul2022,hay2014selecting}. In the project selection task, the meta-level MDP has 
$(max_{obs}-min_{obs}+2)^{N_P \cdot N_C \cdot N_E}$
%$(N_P \cdot N_C)^{(max_{obs}-min_{obs}+1) \cdot N_E+1}$ 
possible belief states and $N_P \cdot N_C \cdot N_E+1$ possible meta-level actions. Our meta-level MDP introduces multiple new intricacies that current metareasoning methods like MGPO \citep{heindrich2025intelligent} aren't equipped to handle, making strategy discovery in this setting especially difficult. Compared to previously formulated meta-level MDPs \citep{callaway2017learning, hay2014selecting, consul2022, heindrich2025intelligent, mehta2022leveraging}, our meta-level description of project selection differs in the following ways: (1) the maximum amount of meta-level actions is constrained with a budget, (2) the project selection task features multiple consultants who differ in both the quality of their advice and the cost of their services, (3) consultants in the project selection task offer discrete estimates of each criterion, 
%on a scale from one to five
requiring that (4) criteria ratings are scaled to allow weighting the criteria according to their importance.

\section{A new metareasoning algorithm for discovering decision strategies for human project selection}

In this section, we present a new metareasoning algorithm based on MGPO \citep{heindrich2025intelligent}. Previous metareasoning methods, including MGPO, are unable to handle some of the intricacies of the project selection problem. Therefore, we developed a new strategy discovery method that overcomes the limitations that prevent MGPO from being applicable to project selection. To reflect the commonalities and innovations, we call our new strategy discovery method MGPS (Meta-Greedy policy for Project Selection; see Algorithm~1). Like MGPO, MGPS approximates the value of computation (VOC) \citep{Russell_Wefald_1991} by myopically estimating the immediate improvement in decision quality. 
%Improving upon MGPO, we adapted the method to allow for multiple consultants who differ in both the quality of their advice and the cost of their services. To make MGPS more suited for naturalistic tasks, we modified the VOC approximation to account for experts giving advice on a scale from one to five. Finally, to account for the different criteria weightings, we implemented a criteria scaling mechanism into the VOC calculation. 
Improving upon MGPO, MGPS calculates the myopic approximation to the VOC in a way that accounts for discrete criteria ratings, criteria scaling, and multiple sources of information with different costs and reliabilities.

\begin{algorithm}
\small
\caption{MGPS VOC calculation for an action $a_{p_i, c_i, e_i}$}\label{alg:voc}
\begin{algorithmic}[1]
\Function{myopic\_voc}{$p_i$, $c_i$, $e_i$, $b$} 
\Statex \hspace{\algorithmicindent} \textit{// Expected reward of selected project}
\State $r_p \gets \mathbb{E}[r^O(p_i)]$\label{alg:rp}
\Statex \hspace{\algorithmicindent} \textit{// Expected reward of best alternative project}
\State $r_{alt} \gets \max_{p_j \in \mathcal{P}-\{p_i\} }\mathbb{E}[r^O(p_j)]$\label{alg:ralt}
\Statex \hspace{\algorithmicindent} \textit{// Current belief of selected criteria}
\State $\mu$, $\sigma \gets b_{p_i, c_i}$
\Statex \hspace{\algorithmicindent} \textit{// Iterate over possible advisor ratings}
\For{$obs$ \textbf{from} $min_{obs}$ \textbf{to} $max_{obs}$} \label{alg:obs_start}
    \Statex \hspace{\algorithmicindent} \textit{// Calculate probability of ratings using the cumulative distribution function}
    \If{$min_{obs} < obs < max_{obs}$} \label{alg:obs_mid}
        \State $p_{obs} \gets \Phi\left(\frac{w_{c_i} \cdot (obs+0.5-\mu)}{\sqrt{(w_{c_i} \cdot \sigma)^2+(w_{c_i} \cdot \sigma_e)^2}}\right) - \Phi\left(\frac{w_{c_i} \cdot (obs-0.5-\mu)}{\sqrt{(w_{c_i} \cdot \sigma)^2+(w_{c_i} \cdot \sigma_e)^2}}\right)$ \label{alg:p_obs}
    \ElsIf{$obs == min_{obs}$} \label{alg:start_p_obs}
        \State $p_{obs} \gets \Phi \left(\frac{w_{c_i} \cdot (obs+0.5-\mu)}{\sqrt{(w_{c_i} \cdot \sigma)^2+(w_{c_i} \cdot \sigma_e)^2}} \right)$ \label{alg:p_obs_min}
    \Else%{$obs == max_{obs}$}
        \State $p_{obs} \gets 1-\Phi\left(\frac{w_{c_i} \cdot (obs-0.5-\mu)}{\sqrt{(w_{c_i} \cdot \sigma)^2+(w_{c_i} \cdot \sigma_e)^2}}\right)$ \label{alg:p_obs_max} 
    \EndIf \label{alg:end_p_obs} \label{alg:obs_end}
    \Statex \hspace{\algorithmicindent} \textit{// Updated belief state after integrating advisor rating}
    \State $\hat{\mu}_{obs} \gets \left(\frac{w_{c_i} \cdot \mu}{(w_{c_i} \cdot \sigma)^2}+\frac{w_{c_i} \cdot obs}{(w_{c_i} \cdot \sigma_e)^2} \right) \cdot \left((w_{c_i} \cdot \sigma)^2+(w_{c_i} \cdot \sigma_e)^2 \right)$ \label{alg:mu_obs}
\EndFor
\Statex \hspace{\algorithmicindent} \textit{// VOC calculation if the investigated project is current best choice}
\If{$r_p > r_{alt}$} \label{alg:if_optimal}
    \State voc $\gets \sum_{obs=min_{obs}}^{max_{obs}} p_{obs}(r_{p_{alt}} + \mu - r_p - \hat{\mu}_{obs}) \cdot \mathds{1}(r_p -\mu + \hat{\mu}_{obs} < r_{alt})$ \label{alg:if_optimal_update}
\Statex \hspace{\algorithmicindent} \textit{// Alternative VOC calculation if another project is the current best choice}
\Else
    \State voc $\gets \sum_{obs=min_{obs}}^{max_{obs}} p_{obs}(r_p + \hat{\mu}_{obs} - \mu - r_{p_{alt}}) \cdot \mathds{1}(r_p - \mu + \hat{\mu}_{obs} > r_{alt})$ \label{alg:not_optimal_update}
\EndIf
\Statex \hspace{\algorithmicindent} \textit{// Final VOC taking costs into account}
\State \Return $(1-w_\lambda)$voc - $w_\lambda\lambda_{e_i}$ \label{alg:cost}
\EndFunction
\vspace{1em}
\Function{mgps}{$b$} \label{alg:mgpo_start}
\Statex \hspace{\algorithmicindent} \textit{// Select computation with highest VOC}
\State $c_{\text{max}} \gets \max_i\textsc{MYOPIC\_VOC}(p_i, c_i, e_i, b)$ 
\If{$c_{\text{max}}>0$}
    \State \Return $c_{\text{max}}$
\Statex \hspace{\algorithmicindent} \textit{// Stop planning if no computation is beneficial}
\ElsIf{$c_{\text{max}}<=0$}
    \State \Return $\perp$
\EndIf
\EndFunction \label{alg:mgpo_end}
\end{algorithmic}
\end{algorithm}

On a high level, MGPS functions by iterating over all possible meta-level actions, approximating their VOC, and iteratively selecting the computation with the highest VOC. When the selected operation requests additional information, the resulting information is used to update the belief state about the evaluated project according to Bayesian learning (Equations~\ref{eq:belief_update_mu}-\ref{eq:belief_update_sigma}). When the selected operation is the termination operation, the policy selects the project with the highest expected utility according to the current belief state. 
%To have a positive VOC, a meta-level action needs to change the belief state in such a way that the project with the highest expected reward changes. This is approximated by evaluating the outcome and likelihood of each possible rating an expert can give.
The method by which MGPS determines when to perform which operation is described in Algorithm~\ref{alg:voc}. 

The VOC of requesting a rating is the difference between the expected utility of receiving the resulting information and its cost. MGPS calculates a myopic approximation of the VOC of asking an expert about a specific criterion of a single project according to Lines~1-21. Concretely, this means that the VOC is approximated by the improvement in expected termination reward that would result from receiving the requested piece of information and then selecting a project compared to selecting a project immediately without obtaining any additional information (Lines~\ref{alg:if_optimal}-\ref{alg:not_optimal_update}).

%Depending on whether the evaluated project is optimal according to the current belief state
The value of requesting a rating depends on the possible posterior belief states that would result from its potential values ($\hat{\mu}_{obs}$) and their probabilities. Crucially, for a computation to have a positive myopic VOC, it must be possible for the resulting rating to change the decision-maker's belief about which project is best. Therefore, the myopic VOC of requesting an expert evaluation chiefly depends on the probability that it will change which project appears to be best. MGPS calculates this probability based on the expected reward of the project that the expert is asked to evaluate ($r_p$) and the expected reward of the best alternative project $r_{alt}$ (Lines~\ref{alg:rp}-\ref{alg:ralt}). If the evaluated project currently has the highest expected reward (see Line~\ref{alg:if_optimal}), the VOC depends on the probability of observing a value that is so low that the currently second-best project would become the most promising option if that value were observed. The potential improvement from selecting a different project is calculated as the difference in expected rewards from the best alternative project $r_{alt}$ and the current best project after integrating the new observation $r_p - \mu + \hat{\mu}_{obs}$ (see Line~\ref{alg:if_optimal_update}). If the evaluated project does not have the highest expected termination reward, the VOC depends on the probability of observing a value high enough to make the evaluated project the most promising option %that the evaluated project has the highest expected termination reward after the belief update 
(see Line~\ref{alg:not_optimal_update}).

The cost of the requested expert $\lambda_{e_i}$ is scaled using a free cost weight parameter $w_\lambda$ and subtracted from the VOC estimate (see Line~\ref{alg:cost}). The cost weight parameter is optimized using Bayesian optimization \citep{mockus2012bayesian} and enables MGPS to balance the number of selected computations in cases where a purely myopic approach might lead to suboptimal results. 

%The VOC of asking an expert about a specific criterion of a specific project depends on the current belief state $(\mu, \sigma)$ of the project's criteria score, the cost $\lambda$ of asking the expert for advice, and the reliability $\sigma_e$ of the expert. 
 %The myopic VOC of requesting an expert evaluation chiefly depends on the probability that it will change which project appears to be best. 
 
%MGPS calculates the expected belief update by iterating over possible observations. For each potential observation, it caluclates the probability that it will occur (Lines~\ref{alg:obs_mid}-\ref{alg:obs_end}) and which belief state it would result in (Line~\ref{alg:mu_obs}). 

To account for the fact that the advisor ratings are discrete rather than continuous, MGPS iterates over the discrete set of ratings that the expert might give and estimates the probability $p_{obs}$ of each rating ($obs$). Expert ratings are modeled using the cumulative distribution function ($\Phi$) of a Normal distribution centered on the current belief state $\mu$ and using the combined standard deviation of the belief $\sigma$ and the expert's reliability $\sigma_e$ (see Line~\ref{alg:p_obs}). The probability of each discrete observation is then estimated by the probability of a sample from the assumed normal distribution falling into an interval around the observation. Concretely, MGPS uses the following intervals: $(-\infty, min_{obs}+0.5]$ for lowest possible value, $(obs-0.5, obs+0.5]$ for intermediate values, and $(max_{obs}-0.5, \infty)$ for the highest possible value. 
%The standard deviation $\sigma_e$ of the likelihood function encodes the expert's reliability, the prior ($\mathcal{N}(\mu,\sigma)$) is the current belief about the project's score on the evaluated criterion, and the weights $w_{c_i}$ convert the criteria into a common currency. 
For each observation, MGPS also calculates the belief update that would result from it ($\hat{\mu}_{obs}$). The simulated update to the belief state $\hat{\mu}$ is computed by applying the belief state update equation (Equation~\ref{eq:belief_update_mu}) to integrate the new observation $obs$ with the prior belief $(\mu, \sigma)$ according to the expert's reliability $\sigma_e$ (see Line~\ref{alg:mu_obs}).

%applying the update equation for a Gaussian likelihood function with standard deviation $\sigma_e$ and a conjugate Gaussian prior (i.e., the current belief).  
%$p(obs) = \Phi(\frac{s*(obs+0.5-\mu)}{\sqrt{(\sigma*s)^2+(\sigma_e*s)^2}}) - \Phi(\frac{s*(obs-0.5-\mu)}{\sqrt{(\sigma*s)^2+(\sigma_e*s)^2}})$ 
%For the highest and lowest possible ratings, we instead calculate $p_{obs}$ using the open interval (see Lines~\ref{alg:p_obs_min} and \ref{alg:p_obs_max}). 
%The updated expected value of the belief state according to an observation $obs$ is then calculated using Bayesian inference to integrate the new observation into the belief state (see Line~\ref{alg:mu_obs}). 
%$\hat{\mu}_{obs}=(\frac{\mu*s}{(\sigma*s)^2}+\frac{obs*s}{(\sigma_e*s)^2})*(\sigma*s)^2+(\sigma_e*s)^2$.
%$\hat{\mu}_{obs}=(\frac{\mu*s}{(\sigma*s)^2}+\frac{obs*s}{(\sigma_e*s)^2})*\sqrt{(\sigma*s)^2+(\sigma_e*s)^2}$. 
% Tau state 
%$(\frac{\mu\tau}{s}+\frac{\text{obs}\tau_e}{s})/(\frac{\tau+\tau_e}{s^2})$

%as well as the expected reward of the evaluated project $r_p$ and the expected reward of the best alternative $r_{p_{alt}}$. If the evaluated project has a higher expected reward than the best alternative, the VOC calculation is: $-\lambda+\sum_{obs=1}^5 p(obs)(r_{p_{alt}} + \mu - r_p - \hat{\mu}_{obs})$. If the investigated project has a lower expected reward than the alternative project, the VOC calculation is: $-\lambda+\sum_{obs=1}^5 p(obs)(r_p + \hat{\mu}_{obs} - \mu - r_{p_{alt}})$.

The full meta-greedy policy (Line~\ref{alg:mgpo_start}) consists in calculating the VOC for all possible meta-level actions and iteratively performing the meta-level action with the highest VOC until no available expert advice has a positive VOC. At that point, the termination action $\perp$ is chosen and the project with the highest expected reward is selected.

\section{Improving human project selection}

%As reviewed above, AI-powered boosting improves human decision-making by combining automatic strategy discovery with intelligent cognitive tutors.  
Having developed a general metareasoning method for discovering resource-rational strategies for human project selection, we now extend it to an intelligent cognitive tutor for teaching people how to select better projects. Our goal is to evaluate whether humans can utilize the strategies discovered by MGPS and to provide a proof of concept for a general AI-powered boosting approach that can be used to improve human decision-making across a wide range of project selection problems. We first introduce a general approach for teaching people the project selection strategies discovered by MGPS, and then apply it to a real-world project selection problem. %Our results demonstrate that people who are taught the strategy discovered by MGPS perform significantly better. %To achieve this, we developed a cognitive tutor that helps people learn how to select better projects by teaching them an efficient project selection strategy discovered by our new metareasoning method. 

\subsection{MGPS Tutor: An intelligent tutor for teaching people how to select better projects}
\begin{figure}[ht]
  \centering
  \includegraphics[width=0.9\textwidth]{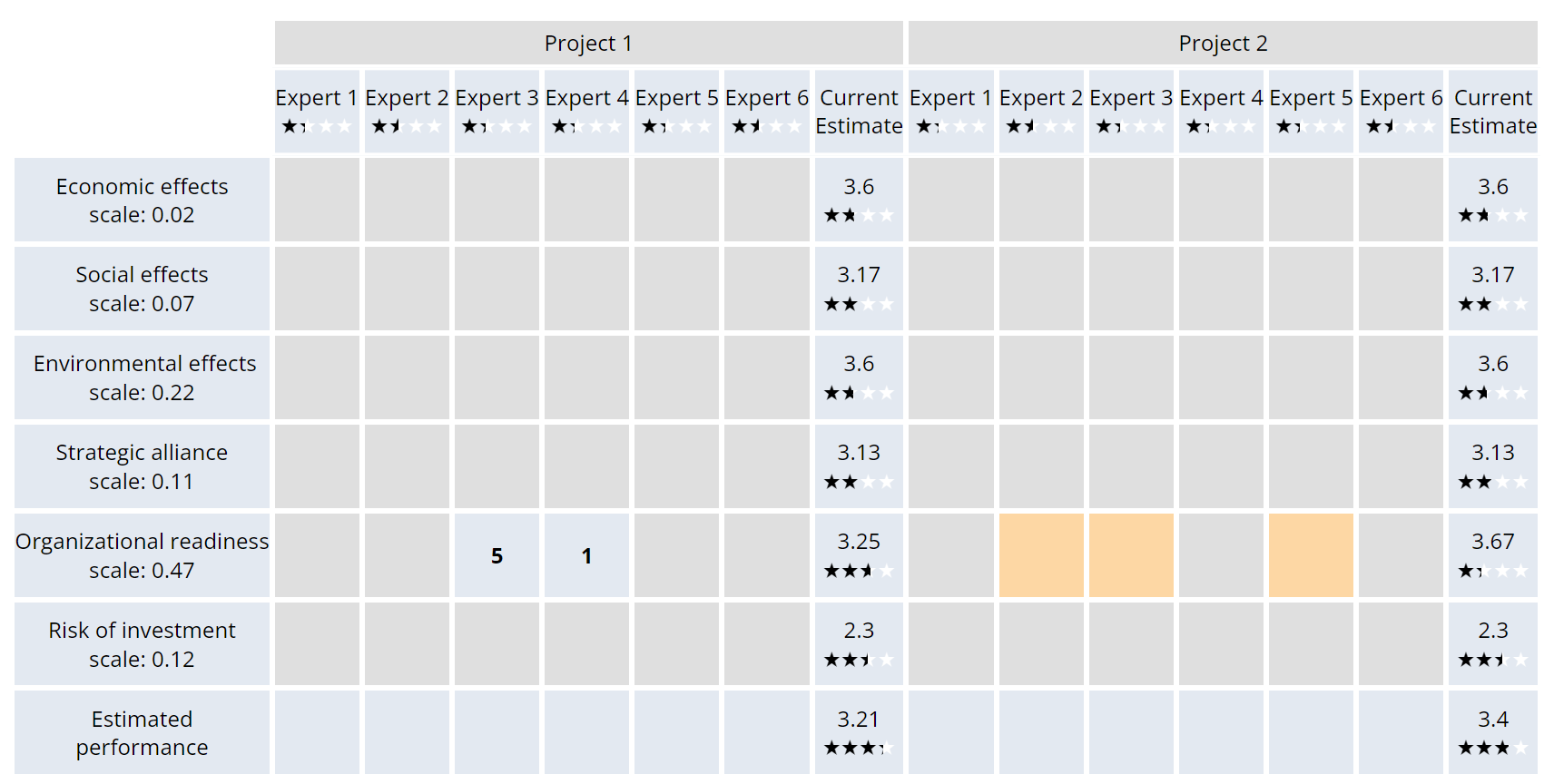}
  \caption{Example of the MGPS tutor offering a choice between requesting information from three different experts (highlighted in orange) in the simplified training task of deciding between two project alternatives.}
  \label{fig:env_exp}
\end{figure}

Our intelligent tutor for project selection (\textit{MGPS Tutor}) trains people to select the near-optimal decision operations identified by MGPS. To achieve this, it lets people practice on a series of project selection problems and gives them feedback. MPGS Tutor leverages MPGS in two ways: i) to pedagogically construct the set of queries the learner is asked to choose from, and ii) to give the learner feedback on their chosen query.

Building on the choice tutor by \citep{heindrich2025intelligent}, our tutor repeatedly asks the learner to choose from a pedagogically chosen set of decision operations (see Figure~\ref{fig:env_exp}) that always includes the query that MGPS would have performed. Moreover, it leverages MGPS's VOC calculation (Algorithm~\ref{alg:voc}) to score the chosen query, and then provides binary feedback on its quality. If learners select a suboptimal expert, project, or criterion, they receive feedback indicating the correct choice and have to wait for a short time. The unpleasantness of having to wait serves as a penalty \citep{Callaway2021Leveraging}. 
Otherwise, they receive positive reinforcement and the next choice is displayed. To receive positive reinforcement, the learner must select a query whose VOC is sufficiently close, as determined by a tolerance parameter $t$, to the VOC of the optimal action. We set the tolerance to $t=0.001$.

Our tutor teaches the strategy discovered by MGPS using a novel sophisticated training schedule, which fosters learning by incrementally increasing the complexity of the training task. This learning methodology is also known as shaping \citep{skinner1953shaping}, and has been successfully applied to teach decision strategies to humans \citep{heindrich2025intelligent}. Our training schedule varies the numbers of projects, how many different expert assessments learners have to choose between, and the specific types of expert assessments offered as choices. In total, our tutor teaches the discovered project selection strategy using ten training trials. The first seven training trials use a smaller version of the project selection task with only two projects, while the last three trials use the full environment with five projects. The number of choices gradually increases throughout training from 1 in the first training trial to 9 in the last three training trials. %, starting with a trial where the choice set includes only a single query
%, four trials in which learners choose between two meta-level actions, two trials with choices between six meta-level actions, 
%, and increasing the number of choices to up to nine meta-level actions for the final three trials. 
%The tutor varies the types of choices across trials, thereby progressing from teaching the learner to choose between experts to teaching them how to choose between criteria to teaching them to choose both jointly. To achieve this, the tutor initially offers learners a single-choice trial that requires them to select the optimal query. Then, the tutor offers three choices that focus on different criteria within the same project for two trials. The following two trials still offer three choices, but focus on offering choices between different experts within the same project. The remaining five trials combine both types of highlights, offering learners a complex decision problem between different experts and criteria, while also increasing the overall number of choices. While the tutor primarily offers choices within the same project, later trials with higher a number of choices sometimes feature queries about different projects. The optimal choice is always included in the choice set. 
The tutor varies the types of choices across trials. After an initial trial with only a single choice, the tutor offers choices that focus on different criteria within the same project for two trials. Then, the tutor offers choices that focus on different experts within the same project for two trials. The remaining trials combine both types of highlights while sometimes also featuring queries about different projects and also increasing the overall number of choices.

\subsection{Evaluating the effectiveness of MGPS Tutor in a training experiment}

To evaluate if AI-powered boosting can improve human project selection, we tested the MPGS tutor in a training experiment. We tested if people trained by MPGS tutor learn more resource-rational project selection strategies. To make our assessment task as naturalistic as possible, we modelled it on a real project selection problem that was faced by an Iranian financial institution \citep{Khalili-Damghani_Sadi-Nezhad_2013}. We will first describe how we modeled this real-world problem, and then the training experiment. 

\paragraph{A project-selection problem from the real world}

\citet{Khalili-Damghani_Sadi-Nezhad_2013} worked on the real-world problem of helping a financial institution select between five potential projects with an eye to sustainability. Each project was evaluated by six advisors, who assigned scores from one to five on six different criteria. For our model of the task, we use the same number of experts, criteria, and projects, and the same criteria weights as the financial institution. The remaining parameters of the meta-level MDP were estimated as follows. We initialized the beliefs about the project's attributes by calculating the mean and the standard deviation of all expert ratings for each criterion according to \citep{Khalili-Damghani_Sadi-Nezhad_2013}. We estimated the reliability of each expert by calculating the standard deviation from the average distance of their ratings from the average rating of all other experts, weighted by the number of occurrences of each guess. We estimated the cost parameter $\lambda$ of the meta-level MDP by $\lambda = \frac{cost}{stakes} \cdot r(\perp)$ to align the meta-level MDP's cost-reward ratio to its real-world equivalent. Using the expected termination reward of the environment $r(\perp)=3.4$ and rough estimates for the stakes $stakes=\$10000000$ and expert costs $cost=\$5000$, this led to $\lambda=0.002$. While \citet{Khalili-Damghani_Sadi-Nezhad_2013} assumed all expert ratings are available for free, in the real world this is rarely the case. To make our test case more representative, we assumed that advisor evaluations are available on-request for a consulting fee. To capture that real-world decisions often have deadlines that limit how much information can be gathered, we set the maximum number of sequentially requested expert consultations to $5$.

\paragraph{Methods of the experiment} \label{sec:methods_exp}
We recruited 301 participants for an online training experiment on \href{https://www.prolific.co/}{Prolific}. The average participant age was 29 years, and 148 participants identified as female. Participants were paid £3.50 for completing the experiment, plus an average bonus of £0.50. The median duration of the experiment was 22 minutes, resulting in a median pay of £10.9 per hour. Our experiment was preregistered on \href{https://aspredicted.org/YNY_H3K}{AsPredicted} and approved by the ethics commission of 
the Medical Faculty of the University of Tübingen 
under IRB protocol number 667/2018BO2. 

Each participant was randomly assigned to one of three conditions: (1) the \textit{No tutor} condition, in which participants did not receive any feedback and were free to discover efficient strategies on their own; (2) the \textit{MGPS tutor} condition, in which participants practiced with our cognitive tutor that provided feedback on the resource-rationality score MGPS assigns to the selected planning actions; and (3) the \textit{Dummy tutor} condition, an additional control condition in which participants practiced with a dummy tutor comparable to the MGPS tutor, albeit with randomized feedback on which planning actions are correct. All participants practiced their planning strategy in 10 training trials and were then evaluated across 10 test trials. For each trial, a new instance of the project selection environment was randomly generated by first sampling the project's criteria scores from the initial belief state and then generating each expert's ratings based on their reliability $\sigma_e$ using a Gaussian distribution centered at the criteria score.

We evaluated the participants' performance using two measures: their \textit{RR-score} and \textit{click agreement} \citep{heindrich2025intelligent}. \textit{RR-score's} are normalized by subtracting the average reward of a random baseline algorithm and dividing by the participant scores' standard deviation. The random baseline algorithm is defined as the policy that chooses meta-level actions at random until the maximum number of decision operations is reached. \textit{Click agreement} measures, how well participants learned to follow the near-optimal strategy discovered by our method. Specifically, we computed for each participant, which proportion of their information requests matched the action taken by the approximately resource-rational strategy discovered by MGPS.

\paragraph{Experiment results}

\begin{table}
  \centering
  \begin{tabular}{lrr}
    \toprule
    % CI 95 results
    Condition & RR-score & Click Agreement \\
    \midrule
     \textbf{MGPS Tutor} & $0.3256  \pm 0.0609$ &$ 0.4271 \pm 0.0201 $ \\
    No tutor & $-0.0227 \pm 0.0622 $ & $0.2521  \pm 0.0171$ \\
   Dummy tutor & $0.0225 \pm 0.0612 $& $0.2664 \pm 0.0159 $\\
    \bottomrule
  \end{tabular}
  \caption{Results of the human training experiment. Per condition, the normalized mean resource-rationality score and the mean click agreement are reported. For both measures, we also report the 95\% confidence interval under the Gaussian assumption ($\pm 1.96$ standard errors). }
  \label{table:exp}
\end{table}

Table~\ref{table:exp} shows the results of the experiment. To determine whether the condition of participants had a significant effect on the RR-score and click agreement, we used an ANOVA analysis with Box approximation \citep{box1954some}. The ANOVA revealed a significant effect of condition on both RR-score ($F(1.99, 293.57)=10.48$, $p<.0001$) and click agreement ($F(1.99, 291.48)=15.5$, $p<.0001$). We further compared the performance of participants in the \emph{MGPS tutor} condition to participants in the two control conditions with post hoc ANOVA-type statistics and used Cohen's d \citep{cohen2013statistical} to evaluate the size of the effects. The post hoc tests revealed that participants in the \emph{MGPS tutor} condition achieved a significantly higher RR-score than participants in the \emph{No tutor} ($F(1)=17.88$, $p<.0001$, $d=.35$) and \emph{Dummy tutor} ($F(1)=13.4$, $p=.0002$, $d=.31$) conditions. Additionally, participants in the \emph{MGPS tutor} reached a higher click agreement with our pre-computed near-optimal strategy than participants in the \emph{No tutor} ($F(1)=25.08$, $p<.0001$, $d=.58$) and \emph{Dummy tutor} ($F(1)=19.3$, $p<.0001$, $d=.56$) conditions.

When evaluated on the same test trials and normalizing against the baseline reward and the standard deviation of the experiment results, MGPS achieves a mean reward of $1.17$, demonstrating that MGPS discovers more resource-rational strategies than participants across all conditions. Although participants in the \textit{MGPS tutor} condition performed significantly the better than participants in the other conditions, they did not learn to follow the strategy taught by the tutor exactly. Participants in the other conditions only discovered strategies with a similar \textit{RR-score} to the random baseline strategy, with participants in the \textit{No tutor} condition performing even worse than the random baseline strategy, and participants in the \textit{Dummy tutor} condition outperforming the random baseline only by a small margin.

\begin{figure}[ht]
  \centering
  \includegraphics[width=0.9\textwidth]{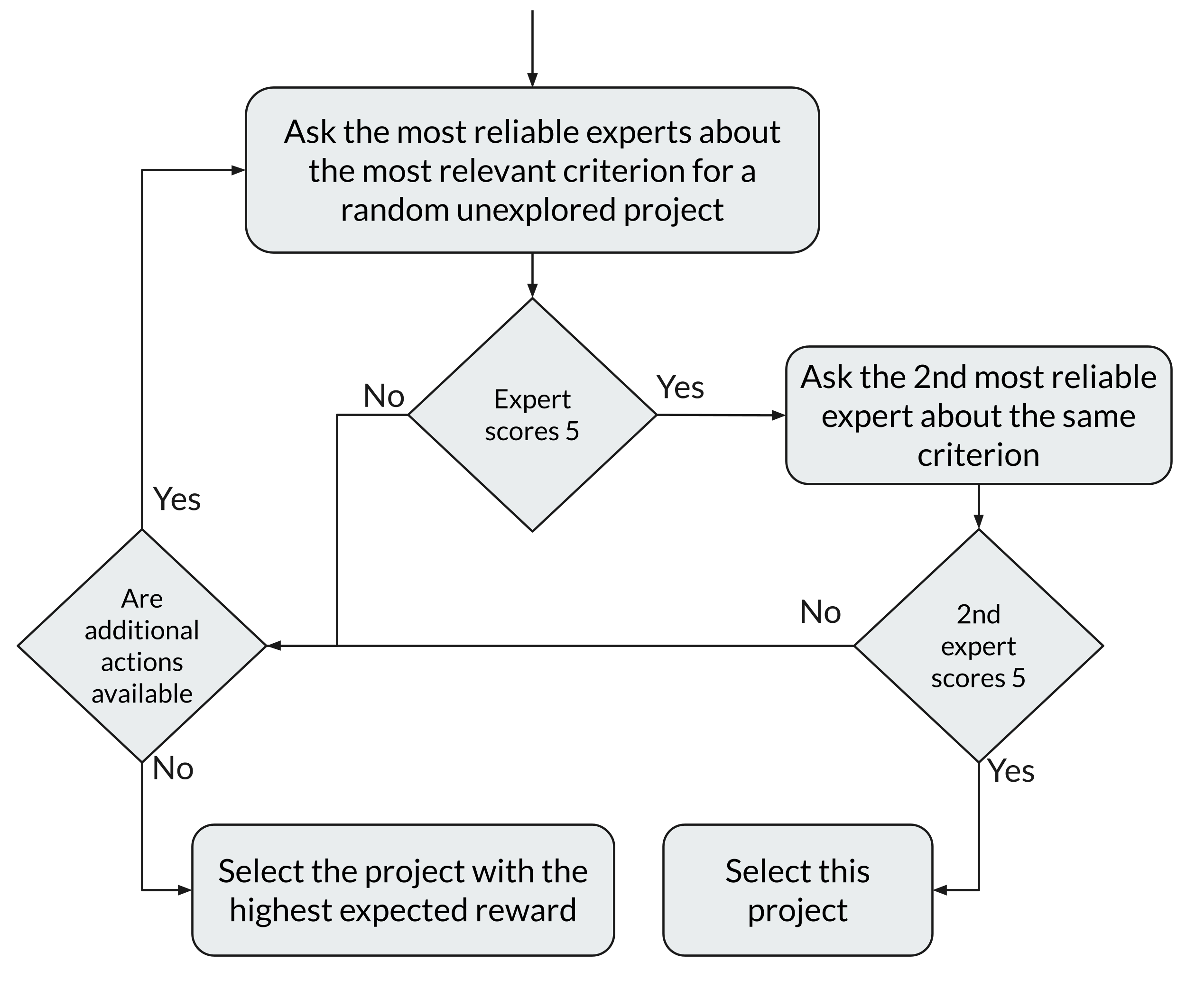}
  \caption{Flowchart explaining the strategy discovered by MGPS.}
  \label{fig:strategy}
\end{figure}

\paragraph{What strategy did the MGPS Tutor teach?}
To understand what participants in the MGPS Tutor condition were taught, we inspected the strategy discovered by MGPS and visualized it as a flowchart in Figure~\ref{fig:strategy}. We found that this strategy systematically asks the most reliable experts ($e_2$ and $e_6$) to evaluate projects on the criterion with the highest decision weight ($c_5$). It starts by asking expert $e_2$ to rate criterion $c_5$ for a randomly selected project. If that rating is below the maximum score, the same expert is immediately asked to rate the same criterion for a different project. If expert $e_2$ gives project $p_i$ the highest possible score on criterion $c_5$ ($max_{obs}=5$), the strategy requests a second opinion about it from expert $e_6$.
If expert $e_6$ also evaluates the criterion's value as $5$, the decision process ends and project $p_i$ is selected. If expert $e_6$ provides a rating below $5$, expert $e_2$ is asked about a different project. This process is repeated until a project in which both experts $e_2$ and $e_6$ estimate $c_5=5$ is found, or the maximum number of decision operations ($5$) is reached. It is important to note that while this decision strategy is relatively easy to understand, describe, and execute, the task of discovering this specific strategy as more resource-rational than the vast number of possible other strategies is considerably more difficult.

\paragraph{What aspects of the strategy did participants learn?}
We further analyzed which aspects of MGPS' strategy participants learned from the intelligent tutor. Specifically, we investigated in which proportion of the test trials participants managed to (1) start planning with an optimal action and (2) correctly decide whether to continue investigating the same project or switch to requesting information about an alternative project. 
MGPS' strategy starts by requesting information about criterion $c_5$ from one of the two most reliable experts. Participants in the \textit{MGPS tutor} condition did so $56\%$ of the time, whereas participants in the \textit{No tutor} condition did so only $33\%$ of the time. Looking into why many failed to request this information, we found that more participants learned to request information from the most reliable experts ($73\%$ for participants in the \textit{MGPS tutor} condition and $43\%$ for participants in the \textit{No tutor} condition) than to request information about the most important evaluation criteria ($64\%$ for participants in the \textit{MGPS tutor} condition and $43\%$ for participants in the \textit{No tutor} condition).

A second important aspect of MGPS's strategy is how it responds to the information revealed. Excluding trials in which participants' first expert request did not match an action MGPS identifies as optimal, we analyzed whether participants either correctly continued to evaluate the current project (in case of revealing an expert guess of $5$), or correctly switched to evaluating an alternative project (when revealing an expert guess smaller than $5$). Participants in the \textit{MGPS tutor} condition ($65\%$ of test trials) and participants in the \textit{No tutor} condition ($65\%$ of test trials) both learned to accurately switch projects when appropriate. Repeatedly evaluating the current project proved to be the most difficult component of the discovered strategy to learn, with participants in the only \textit{MGPS tutor} condition following this aspect of the strategy in $11\%$ of test trials and participants of the \textit{No tutor} condition following the strategy in only $7\%$ of test trials. 

\section{Performance evaluation}

The results reported in the previous section show that MPGS can discover project selection strategies that are more effective than the strategies people discover on their own. But how does its performance compare to other strategy discovery algorithms? To answer this question, we evaluated MGPS on a computational benchmark. We chose PO-UCT \citep{silver2010monte} for comparisons because it is an established baseline for metareasoning algorithms in partially observable environments \citep{heindrich2025intelligent} and the more specialized MGPO algorithm is not applicable to project selection. PO-UCT utilizes Monte Carlo tree search to simulate the effects of different actions, resulting in more accurate results with increased computation time, making it a useful baseline for MGPS's computational efficiency and performance. 
%Due to the computational complexity of the meta-level MDP, we didn't expect PO-UCT to find the optimal solution in our benchmark experiments. However, we believe it to be a useful baseline for MGPS's computational efficiency and performance.

\paragraph{Method}

We evaluated the effectiveness of our method in the project selection task by comparing it against PO-UCT \citep{silver2010monte} with different numbers of steps. All methods were evaluated across the same 5000 randomly generated instances of the project selection environment. 

Our main performance measure was the resource-rationality score (\textit{RR-Score} defined in Equation~\ref{eq:rr}). To highlight the achieved improvements over a baseline algorithm that performs random meta-level actions, we normalized the reported \textit{RR-scores}. Specifically, we applied a z-score transformation, subtracting the average reward of the random baseline algorithm (see Section~\ref{sec:methods_exp}) from the \textit{RR-Scores} and dividing by the evaluated algorithm's \textit{RR-Scores}' standard deviation. We analyze the differences in \textit{RR-Scores} with an ANOVA and evaluate the size of statistical effects with Cohen's d \citep{cohen2013statistical}. Additionally, we compare the computational efficiency of the different methods, which is crucial for being able to provide real-time feedback in our cognitive tutor. 

\paragraph{Results}

\begin{table}
  \centering
  \begin{tabular}{lrr}
    \toprule

    % CI95 scores
    Algorithm & RR-score & Runtime (s) \\
    \midrule
    \textbf{MGPS} & $0.9942 \pm 0.0234$  & $0.9079 \pm 0.0052$  \\
    PO-UCT (10 steps) & $-0.4344 \pm 0.0106$  & $0.0175 \pm 0.0004$  \\
    PO-UCT (100 steps) & $0.7309 \pm 0.0302$  & $0.1972 \pm 0.0008$  \\
    PO-UCT (1000 steps) & $0.8681 \pm 0.0256$ 	 & $2.3567 \pm 0.0028$  \\
    PO-UCT (5000 steps) & $0.9054 \pm 0.0232$  & $10.8913 \pm 0.0173$  \\
    \bottomrule
  \end{tabular}
   \caption{Results of the performance evaluation. For each algorithm, we report the averge normalized resource-rationality score (\textit{RR-Scores}) and the runtime per decision problem. For both measures, we also report the 95\% confidence interval under the Gaussian assumption ($\pm 1.96$ standard errors).}
   \label{table:sim}
\end{table}

As shown in Table~\ref{table:sim}, MGPS outperformed all tested versions of PO-UCT and the random baseline strategy (as the \textit{RR-scores} are normalized by subtracting the mean \textit{RR-score} of the random baseline, the random baseline strategy itself has a normalized \textit{RR-score} of $0$). 
An ANOVA revealed significant differences in the \textit{RR-scores} of the strategies discovered by the different methods ($F(4, 24995)=2447$, $p<.0001$). Hukey-HSD post-hoc comparisons indicated that the strategies discovered by MGPS are significantly more resource-rational than the strategies discovered by PO-UCT with 10 steps ($p<.0001$, $d=2.18$), 100 steps ($p<.0001$, $d=.27$), 1000 steps ($p<.0001$, $d=.14$), or 5000 steps ($p<.0001$, $d=.11$). While MGPS achieves significantly higher \textit{RR-scores} than all PO-UCT variants, the size of the effect decreases from a very large effect to a small effect when increasing PO-UCT's computational budget sufficiently. We therefore expect that PO-UCT with a much more than 5000 steps would ultimately achieve comparable \textit{RR-scores} to MGPO, albeit at a much higher computational cost.
Moreover, MGPS was substantially faster than PO-UCT with a computational budget of 1000 steps or more, which is when PO-UCT's performance starts to approach that of MGPS. With a computational budget of 100 steps or fewer, PO-UCT is faster than MGPS. However, such a small computational budget is not enough for PO-UCT to discover strategies with a \textit{RR-score} anywhere near that of the strategy discovered by MGPS.
%Even when giving PO-UCT a computational budget 10 times higher than our method requires, PO-UCT does not achieve similar \textit{RR-scores} to our method. 
Critically, the high amount of computation required for PO-UCT to achieve an approximately similar level of resource-rationality would render PO-UCT unusable for a cognitive tutor that computes feedback in real time.

\section{Conclusion}

%People's decision-making is prone to systematic errors \citep{kahneman1982judgment}, and although people are happy to delegate some decisions, most CEOs are unlikely to let AI decide which projects their company should pursue. Moreover, people are reluctant to use normative project selection procedures because they tend to be more tedious \citep{schmidt_recent_1992, liu_evaluation_2017, kornfeld_selection_2013}. Motivated by people's insistence on making their own decisions with simple heuristics, we leveraged AI to discover and teach prescriptive decision strategies adapted to human cognition that perform substantially better than people's intuitive strategies but are nevertheless simple enough that people use them.
%this sometimes problematic combination, we introduced a method that leverages AI to boost people's competency to select good projects.
This article presented a first proof-of-concept demonstration that AI can be applied to discover and teach cognitive strategies enabling people to make better decisions in the real world. 

% Successes
To this end, we introduced a metareasoning method for leveraging AI to discover near-optimal decision strategies for human project selection. Modeling project selection through the lens of resource rationality allowed us to formulate a mathematically precise criterion for the quality of decision strategies for human project selection. % and objectively assessing the quality of decision strategies.
We further developed an efficient automatic strategy discovery algorithm that automatically discovers efficient strategies for human project selection. Our algorithm discovered decision strategies that are much more resource-rational than the strategies humans discovered on their own and the strategies discovered by a general-purpose algorithm for solving POMDPs (PO-UCT). Using the decision strategies discovered by our algorithm, we created a cognitive tutor that uses a shaping schedule and metacognitive feedback to teach the strategies to humans. In the training experiment, our cognitive tutor fostered significant improvements in participants' resource rationality.

Prior work has extended strategy discovery to partially observable environments, a critical advance for utilizing strategy discovery to improve human decision-making in the real world \citep{heindrich2025intelligent}. While the intelligent tutor in prior work proved effective at modeling partially observable problems and improving human decision-making, it shared a key limitation with previous work in the field: it was evaluated on abstract and simplified toy problems. In this article, we demonstrated a viable pathway to bringing strategy discovery into the real world: using real-world data to construct metalevel MDPs of real problems, solving the metalevel MDP with strategy discovery methods, and teaching the discovered strategies to humans with intelligent tutors.

This approach might make it possible to extend the curricula of intelligent tutoring systems beyond the procedural skills that instructional designers are readily able to articulate. Our demonstration that this approach is applicable to a decision-making skill commonly taught in MBA programs could become a milestone on the path to integrating automatic strategy discovery into educational applications of intelligent tutoring systems.
%This constitutes an important step towards automatically designing optimal learning curricula for cognitive skills without relying on human intuitions. 

% Limitations and future work
A main limitation of our method is that it is unknown how precisely the environment parameters need to be estimated to construct the metareasoning task. With a highly accurate environment model, it is more likely that people will be able to directly apply the learned strategies to analogous situations in the real world, while a more abstract representation of the real-world task would require significant transfer of the learned strategy. While limited evidence for strategy transfer exists \citep{callaway2022Rational}, we believe modeling real-world problems as accurately as possible is the most likely path towards real-world impact. This can prove especially problematic when there isn't much data on past decisions. Future work could investigate this issue with a sensitivity analysis and potentially address it by extending MGPS with a Bayesian inference approach to estimate uncertain parameters of the environment \citep{mehta2022leveraging}. This extension would make MGPS robust to uncertainties in the underlying distributions of evaluation criteria or expert reliability, and make it possible to discover resource-rational strategies even when data to estimate environment parameters is limited. 

Another simplification of the environment model is that all observations are generated by retrieving a noisy sample of the true reward distribution, which is likely overly optimistic of real-world situations. For example, experts in the project evaluation task might have biases that lead to systematic overvaluation of specific criteria. This limitation could be addressed by integrating a model of people's biases, such as utility weighted sampling \citep{lieder2015utility}. 

Our method, MGPS, is limited by the assumption that each computation is independent from all other computations. Real-world situations might involve more complex, interdependent planning steps. While it is possible to integrate this into the metalevel-MDP, MGPS's myopic approximation could fail in more complicated scenarios with interdependent criteria. Such challenges could be addressed by solving meta-level MDPs with methods from deep reinforcement learning, for example, by utilizing AlphaZero \citep{alphazero}.

Similarly, while our cognitive tutor proved to be effective in teaching humans critical aspects of MGPS' discovered strategies, we are looking forward to future pedagogical work that improves and further evaluates the mechanisms by which the cognitive tutor teaches strategies. An especially exciting direction could be to expand the tutor with automatically generated natural language descriptions for how to make project selection decisions, such as using AI-Interpret \citep{skirzynski2021automatic} or large language models.

While our implementation of project selection demonstrated the viability of improving human decision-making in naturalistic tasks, we did not evaluate our method directly in the real world. 
The most critical direction of future work is therefore to assess the methodology in a real-world field test. We believe our method to be generally applicable to a wide range of real-world decision-making problems that follow the general structure of being able to improve the quality of a decision (i.e. being able to choose a better project after learning additional information) at the cost of expending additional planning effort (i.e. the cost of requesting an expert opinion). Building upon our project selection model, relevant real-world decisions could be a company deciding which research project to invest in or a charity deciding which initiatives maximize impact and cost-effectiveness. 

% Conclusion
Our results indicate that it is possible to use resource-rational analysis combined with automatic strategy discovery to improve human decision-making in realistic scenarios. While our main goal was to provide a proof-of-concept demonstration for the general approach, project selection is an important practical skill that is commonly taught in MBA programs. Future work could leverage automatic strategy discovery methods to design of intelligent tutoring systems for training executive and managerial decision making. The resulting educational technologies could be useful for equipping future leaders with more rational, yet practical, project selection strategies. More generally, we view our results as demonstrating the feasibility of (1) modeling good real-world decision-making within the rational metreasoning framework, (2) discovering efficient cognitive strategies for real-world problems, and (3) improving human decision-making by teaching them the discovered strategies. We are optimistic that this general approach can ultimately be applied to teaching cognitive skills, including mathematical problem-solving, decision-making, and computational thinking, in educational settings ranging from high school to executive MBA programs.

\subsection*{Acknowledgements}

This project was funded by grant number CyVy-RF-2019-02 from the Cyber Valley Research Fund.\\

\noindent
The authors thank the International Max Planck Research School for Intelligent Systems (IMPRS-IS) for supporting Lovis Heindrich.\\

\noindent
The code of the strategy discovery method as well as data and analysis for the simulation and training experiment are available online: \url{https://github.com/lovis-heindrich/MGPS-project-selection/}.

\bibliography{refs}

\begin{thebibliography}{54}
\providecommand{\natexlab}[1]{#1}
\providecommand{\url}[1]{\texttt{#1}}
\expandafter\ifx\csname urlstyle\endcsname\relax
  \providecommand{\doi}[1]{doi: #1}\else
  \providecommand{\doi}{doi: \begingroup \urlstyle{rm}\Url}\fi

\bibitem[Abdel-Basset et~al.(2019)Abdel-Basset, Atef, and Smarandache]{abdel2019hybrid}
M.~Abdel-Basset, A.~Atef, and F.~Smarandache.
\newblock A hybrid neutrosophic multiple criteria group decision making approach for project selection.
\newblock \emph{Cognitive Systems Research}, 57:\penalty0 216--227, 2019.

\bibitem[Aleven et~al.(2006)Aleven, McLaren, Roll, and Koedinger]{aleven2006toward}
V.~Aleven, B.~M. McLaren, I.~Roll, and K.~R. Koedinger.
\newblock Toward meta-cognitive tutoring: {A} model of help seeking with a cognitive tutor.
\newblock \emph{Int. J. Artif. Intell. Educ.}, 16\penalty0 (2):\penalty0 101--128, 2006.
\newblock URL \url{http://content.iospress.com/articles/international-journal-of-artificial-intelligence-in-education/jai16-2-02}.

\bibitem[Aleven et~al.(2009)Aleven, Mclaren, Sewall, and Koedinger]{aleven2009new}
V.~Aleven, B.~M. Mclaren, J.~Sewall, and K.~R. Koedinger.
\newblock A new paradigm for intelligent tutoring systems: Example-tracing tutors.
\newblock \emph{International Journal of Artificial Intelligence in Education}, 19\penalty0 (2):\penalty0 105--154, 2009.

\bibitem[Anderson(1982)]{anderson1982acquisition}
J.~R. Anderson.
\newblock Acquisition of cognitive skill.
\newblock \emph{Psychological review}, 89\penalty0 (4):\penalty0 369, 1982.

\bibitem[Anderson et~al.(1985)Anderson, Boyle, and Reiser]{anderson1985intelligent}
J.~R. Anderson, C.~F. Boyle, and B.~J. Reiser.
\newblock Intelligent tutoring systems.
\newblock \emph{Science}, 228\penalty0 (4698):\penalty0 456--462, 1985.
\newblock \doi{10.1126/science.228.4698.456}.
\newblock URL \url{https://www.science.org/doi/abs/10.1126/science.228.4698.456}.

\bibitem[Becker et~al.(2022)Becker, Skirzy{\'n}ski, van Opheusden, and Lieder]{becker2022boosting}
F.~Becker, J.~Skirzy{\'n}ski, B.~van Opheusden, and F.~Lieder.
\newblock Boosting human decision-making with ai-generated decision aids.
\newblock \emph{Computational Brain \& Behavior}, pages 1--24, 2022.

\bibitem[Bonaccio and Dalal(2006)]{Bonaccio_Dalal_2006}
S.~Bonaccio and R.~S. Dalal.
\newblock Advice taking and decision-making: An integrative literature review, and implications for the organizational sciences.
\newblock \emph{Organizational behavior and human decision processes}, 101\penalty0 (2):\penalty0 127–151, 2006.

\bibitem[Boutilier(2002)]{boutilier2002pomdp}
C.~Boutilier.
\newblock A pomdp formulation of preference elicitation problems.
\newblock In \emph{AAAI/IAAI}, pages 239--246. Edmonton, AB, 2002.

\bibitem[Box et~al.(1954)]{box1954some}
G.~E. Box et~al.
\newblock Some theorems on quadratic forms applied in the study of analysis of variance problems, i. effect of inequality of variance in the one-way classification.
\newblock \emph{The annals of mathematical statistics}, 25\penalty0 (2):\penalty0 290--302, 1954.

\bibitem[Callaway et~al.(2018{\natexlab{a}})Callaway, Gul, Krueger, Griffiths, and Lieder]{callaway2017learning}
F.~Callaway, S.~Gul, P.~M. Krueger, T.~L. Griffiths, and F.~Lieder.
\newblock Learning to select computations.
\newblock \emph{Uncertainty in Artificial Intelligence}, 2018{\natexlab{a}}.

\bibitem[Callaway et~al.(2018{\natexlab{b}})Callaway, Lieder, Das, Gul, Krueger, and Griffiths]{callaway2018resource}
F.~Callaway, F.~Lieder, P.~Das, S.~Gul, P.~M. Krueger, and T.~Griffiths.
\newblock A resource-rational analysis of human planning.
\newblock In \emph{CogSci}, 2018{\natexlab{b}}.

\bibitem[Callaway et~al.(2022{\natexlab{a}})Callaway, Jain, van Opheusden, Das, Iwama, Gul, Krueger, Becker, Grifﬁths, and Lieder]{Callaway2021Leveraging}
F.~Callaway, Y.~R. Jain, B.~van Opheusden, P.~Das, G.~Iwama, S.~Gul, P.~M. Krueger, F.~Becker, T.~L. Grifﬁths, and F.~Lieder.
\newblock Leveraging artificial intelligence to improve people's planning strategies.
\newblock \emph{Proceedings of the National Academy of Sciences of the United States of America}, Mar. 2022{\natexlab{a}}.
\newblock \doi{10.1073/pnas.2117432119}.
\newblock URL \url{https://www.pnas.org/doi/10.1073/pnas.2117432119}.

\bibitem[Callaway et~al.(2022{\natexlab{b}})Callaway, van Opheusden, Gul, Das, Krueger, Griffiths, and Lieder]{callaway2022Rational}
F.~Callaway, B.~van Opheusden, S.~Gul, P.~Das, P.~M. Krueger, T.~L. Griffiths, and F.~Lieder.
\newblock Rational use of cognitive resources in human planning.
\newblock \emph{Nature Human Behaviour}, 6\penalty0 (8):\penalty0 1112--1125, 2022{\natexlab{b}}.
\newblock \doi{10.1038/s41562-022-01332-8}.

\bibitem[Carazo et~al.(2012)Carazo, Contreras, Gómez, and Pérez]{carazo_project_2012}
A.~F. Carazo, I.~Contreras, T.~Gómez, and F.~Pérez.
\newblock A project portfolio selection problem in a group decision-making context.
\newblock \emph{Journal of Industrial \& Management Optimization}, 8\penalty0 (1):\penalty0 243, 2012.
\newblock Publisher: American Institute of Mathematical Sciences.

\bibitem[Chi and VanLehn(2010)]{chi2010meta}
M.~Chi and K.~VanLehn.
\newblock Meta-cognitive strategy instruction in intelligent tutoring systems: how, when, and why.
\newblock \emph{Journal of Educational Technology \& Society}, 13\penalty0 (1):\penalty0 25--39, 2010.

\bibitem[Cohen(2013)]{cohen2013statistical}
J.~Cohen.
\newblock \emph{Statistical power analysis for the behavioral sciences}.
\newblock Academic press, 2013.

\bibitem[Coldrick et~al.(2002)Coldrick, Lawson, Ivey, and Lockwood]{Coldrick_Lawson_Ivey_Lockwood_2002}
S.~Coldrick, C.~Lawson, P.~Ivey, and C.~Lockwood.
\newblock A decision framework for r\&d project selection.
\newblock In \emph{IEEE International Engineering Management Conference}, volume~1, page 413–418. IEEE, 2002.

\bibitem[Collins et~al.(1991)Collins, Brown, Holum, et~al.]{collins1991cognitive}
A.~Collins, J.~S. Brown, A.~Holum, et~al.
\newblock Cognitive apprenticeship: Making thinking visible.
\newblock \emph{American educator}, 15\penalty0 (3):\penalty0 6--11, 1991.

\bibitem[Consul et~al.(2022)Consul, Heindrich, Stojcheski, and Lieder]{consul2022}
S.~Consul, L.~Heindrich, J.~Stojcheski, and F.~Lieder.
\newblock Improving human decision-making by discovering efficient strategies for hierarchical planning.
\newblock \emph{Computational Brain \& Behavior}, 5\penalty0 (2):\penalty0 185--216, 2022.

\bibitem[de~Souza et~al.(2021)de~Souza, dos Santos, Soma, and da~Silva]{de_souza_mcdm-based_2021}
D.~G.~B. de~Souza, E.~A. dos Santos, N.~Y. Soma, and C.~E.~S. da~Silva.
\newblock {MCDM}-{Based} {R}\&{D} {Project} {Selection}: {A} {Systematic} {Literature} {Review}.
\newblock \emph{Sustainability}, 13\penalty0 (21):\penalty0 11626, 2021.
\newblock Publisher: MDPI.

\bibitem[Doshi and Roy(2008)]{doshi2008permutable}
F.~Doshi and N.~Roy.
\newblock The permutable pomdp: fast solutions to pomdps for preference elicitation.
\newblock In \emph{Proceedings of the 7th international joint conference on Autonomous agents and multiagent systems-Volume 1}, pages 493--500, 2008.

\bibitem[Gino(2008)]{gino2008we}
F.~Gino.
\newblock Do we listen to advice just because we paid for it? the impact of advice cost on its use.
\newblock \emph{Organizational Behavior and Human Decision Processes}, 107:\penalty0 234--245, 2008.

\bibitem[Graesser et~al.(2012)Graesser, Conley, and Olney]{graesser2012intelligent}
A.~C. Graesser, M.~W. Conley, and A.~Olney.
\newblock Intelligent tutoring systems.
\newblock \emph{APA educational psychology handbook, Vol 3: Application to learning and teaching.}, pages 451--473, 2012.

\bibitem[Graham and Perin(2007)]{graham2007meta}
S.~Graham and D.~Perin.
\newblock A meta-analysis of writing instruction for adolescent students.
\newblock \emph{Journal of educational psychology}, 99\penalty0 (3):\penalty0 445, 2007.

\bibitem[Guerra and Mellado(2017)]{guerra2017book}
E.~Guerra and G.~Mellado.
\newblock A-book: A feedback-based adaptive system to enhance meta-cognitive skills during reading.
\newblock \emph{Frontiers in Human Neuroscience}, 11:\penalty0 98, 2017.

\bibitem[Hay et~al.(2014)Hay, Russell, Tolpin, and Shimony]{hay2014selecting}
N.~Hay, S.~Russell, D.~Tolpin, and S.~E. Shimony.
\newblock Selecting computations: Theory and applications.
\newblock \emph{arXiv preprint arXiv:1408.2048}, 2014.

\bibitem[Heindrich et~al.(2025)Heindrich, Consul, and Lieder]{heindrich2025intelligent}
L.~Heindrich, S.~Consul, and F.~Lieder.
\newblock An intelligent tutor for planning in large partially observable environments.
\newblock \emph{International Journal of Artificial Intelligence in Education}, pages 1--33, 2025.

\bibitem[Henriksen and Traynor(1999)]{Henriksen_Traynor_1999}
A.~Henriksen and A.~Traynor.
\newblock A practical r\&d project-selection scoring tool.
\newblock \emph{IEEE Transactions on Engineering Management}, 46\penalty0 (2):\penalty0 158–170, 1999.
\newblock \doi{10.1109/17.759144}.

\bibitem[Hertwig and Gr{\"u}ne-Yanoff(2017)]{hertwig2017nudging}
R.~Hertwig and T.~Gr{\"u}ne-Yanoff.
\newblock Nudging and boosting: Steering or empowering good decisions.
\newblock \emph{Perspectives on Psychological Science}, 12\penalty0 (6):\penalty0 973--986, 2017.

\bibitem[Kahneman et~al.(1982)Kahneman, Slovic, Slovic, and Tversky]{kahneman1982judgment}
D.~Kahneman, S.~P. Slovic, P.~Slovic, and A.~Tversky.
\newblock \emph{Judgment under uncertainty: Heuristics and biases}.
\newblock Cambridge university press, 1982.

\bibitem[Khalili-Damghani and Sadi-Nezhad(2013)]{Khalili-Damghani_Sadi-Nezhad_2013}
K.~Khalili-Damghani and S.~Sadi-Nezhad.
\newblock A hybrid fuzzy multiple criteria group decision making approach for sustainable project selection.
\newblock \emph{Applied Soft Computing}, 13\penalty0 (1):\penalty0 339–352, 2013.

\bibitem[Koedinger et~al.(1997)Koedinger, Anderson, Hadley, and Mark]{koedinger1997intelligent}
K.~R. Koedinger, J.~R. Anderson, W.~H. Hadley, and M.~A. Mark.
\newblock Intelligent tutoring goes to school in the big city.
\newblock \emph{International Journal of Artificial Intelligence in Education}, 8:\penalty0 30--43, 1997.

\bibitem[Kornfeld and Kara(2013)]{kornfeld_selection_2013}
B.~Kornfeld and S.~Kara.
\newblock Selection of {Lean} and {Six} {Sigma} projects in industry.
\newblock \emph{International Journal of Lean Six Sigma}, 2013.
\newblock Publisher: Emerald Group Publishing Limited.

\bibitem[Krueger et~al.(2024)Krueger, Callaway, Gul, Griffiths, and Lieder]{krueger2024identifying}
P.~M. Krueger, F.~Callaway, S.~Gul, T.~L. Griffiths, and F.~Lieder.
\newblock Identifying resource-rational heuristics for risky choice.
\newblock \emph{Psychological Review}, 131\penalty0 (4):\penalty0 905, 2024.

\bibitem[Lieder and Griffiths(2020)]{LiederGriffiths2020}
F.~Lieder and T.~L. Griffiths.
\newblock Resource-rational analysis: understanding human cognition as the optimal use of limited computational resources.
\newblock \emph{Behavioral and Brain Sciences}, 43, 2020.

\bibitem[Lieder et~al.(2015)Lieder, Griffiths, and Hsu]{lieder2015utility}
F.~Lieder, T.~L. Griffiths, and M.~Hsu.
\newblock Utility-weighted sampling in decisions from experience.
\newblock In \emph{The 2nd Multidisciplinary Conference on Reinforcement Learning and Decision Making}, 2015.

\bibitem[Lieder et~al.(2019)Lieder, Callaway, Jain, Krueger, Das, Gul, and Griffiths]{CognitiveTutorsRLDM}
F.~Lieder, F.~Callaway, Y.~Jain, P.~Krueger, P.~Das, S.~Gul, and T.~Griffiths.
\newblock A cognitive tutor for helping people overcome present bias.
\newblock In \emph{{RLDM 2019}}, 2019.

\bibitem[Liu et~al.(2017)Liu, Zhu, Chen, Xu, and Yang]{liu_evaluation_2017}
F.~Liu, W.-d. Zhu, Y.-w. Chen, D.-l. Xu, and J.-b. Yang.
\newblock Evaluation, ranking and selection of {R}\&{D} projects by multiple experts: an evidential reasoning rule based approach.
\newblock \emph{Scientometrics}, 111\penalty0 (3):\penalty0 1501--1519, 2017.
\newblock Publisher: Springer.

\bibitem[Mehta et~al.(2022)Mehta, Jain, Kemtur, Stojcheski, Consul, To{\v{s}}i{\'c}, and Lieder]{mehta2022leveraging}
A.~Mehta, Y.~R. Jain, A.~Kemtur, J.~Stojcheski, S.~Consul, M.~To{\v{s}}i{\'c}, and F.~Lieder.
\newblock Leveraging machine learning to automatically derive robust decision strategies from imperfect knowledge of the real world.
\newblock \emph{Computational Brain \& Behavior}, 5\penalty0 (3):\penalty0 343--377, 2022.

\bibitem[Mockus(2012)]{mockus2012bayesian}
J.~Mockus.
\newblock \emph{Bayesian approach to global optimization: theory and applications}, volume~37.
\newblock Springer Science \& Business Media, 2012.

\bibitem[Mohagheghi et~al.(2019)Mohagheghi, Mousavi, Antuchevičienė, and Mojtahedi]{mohagheghi_project_2019}
V.~Mohagheghi, S.~M. Mousavi, J.~Antuchevičienė, and M.~Mojtahedi.
\newblock Project portfolio selection problems: a review of models, uncertainty approaches, solution techniques, and case studies.
\newblock \emph{Technological and Economic Development of Economy}, 25\penalty0 (6):\penalty0 1380--1412, 2019.

\bibitem[Olsen et~al.(2019)Olsen, Roepstorff, and Bang]{Olsen_Roepstorff_Bang_2019}
K.~Olsen, A.~Roepstorff, and D.~Bang.
\newblock Knowing whom to learn from: individual differences in metacognition and weighting of social information.
\newblock \emph{PsyArXiv}, 2019.

\bibitem[{\"O}zsoy and Ataman(2009)]{ozsoy2009effect}
G.~{\"O}zsoy and A.~Ataman.
\newblock The effect of metacognitive strategy training on mathematical problem solving achievement.
\newblock \emph{International Electronic Journal of Elementary Education}, 1\penalty0 (2):\penalty0 67--82, 2009.

\bibitem[Ronayne et~al.(2019)Ronayne, Sgroi, et~al.]{Ronayne_Sgroi_others_2019}
D.~Ronayne, D.~Sgroi, et~al.
\newblock \emph{Ignoring good advice}.
\newblock University of Warwick, Centre for Competitive Advantage in the Global …, 2019.

\bibitem[Russell and Wefald(1991)]{Russell_Wefald_1991}
S.~Russell and E.~Wefald.
\newblock Principles of metareasoning.
\newblock \emph{Artificial intelligence}, 49\penalty0 (1–3):\penalty0 361–395, 1991.

\bibitem[Sadi-Nezhad(2017)]{Sadi-Nezhad_2017}
S.~Sadi-Nezhad.
\newblock A state-of-art survey on project selection using mcdm techniques.
\newblock \emph{Journal of Project Management}, 2\penalty0 (1):\penalty0 1–10, 2017.

\bibitem[Schmidt and Freeland(1992)]{schmidt_recent_1992}
R.~L. Schmidt and J.~R. Freeland.
\newblock Recent progress in modeling {R}\&{D} project-selection processes.
\newblock \emph{IEEE Transactions on Engineering Management}, 39\penalty0 (2):\penalty0 189--201, 1992.
\newblock Publisher: IEEE.

\bibitem[Silver and Veness(2010)]{silver2010monte}
D.~Silver and J.~Veness.
\newblock Monte-carlo planning in large pomdps.
\newblock \emph{Advances in neural information processing systems}, 23, 2010.

\bibitem[Silver et~al.(2017)Silver, Hubert, Schrittwieser, Antonoglou, Lai, Guez, Lanctot, Sifre, Kumaran, Graepel, Lillicrap, Simonyan, and Hassabis]{alphazero}
D.~Silver, T.~Hubert, J.~Schrittwieser, I.~Antonoglou, M.~Lai, A.~Guez, M.~Lanctot, L.~Sifre, D.~Kumaran, T.~Graepel, T.~P. Lillicrap, K.~Simonyan, and D.~Hassabis.
\newblock Mastering chess and shogi by self-play with a general reinforcement learning algorithm.
\newblock \emph{CoRR}, abs/1712.01815, 2017.
\newblock URL \url{http://arxiv.org/abs/1712.01815}.

\bibitem[Skinner(1953)]{skinner1953shaping}
B.~Skinner.
\newblock \emph{Shaping and maintaining operant behavior}, pages 91--106.
\newblock Free Press New York, 1953.

\bibitem[Skirzy{\'n}ski et~al.(2021)Skirzy{\'n}ski, Becker, and Lieder]{skirzynski2021automatic}
J.~Skirzy{\'n}ski, F.~Becker, and F.~Lieder.
\newblock Automatic discovery of interpretable planning strategies.
\newblock \emph{Machine Learning}, 110:\penalty0 2641--2683, 2021.

\bibitem[VanLehn(2011)]{vanlehn2011relative}
K.~VanLehn.
\newblock The relative effectiveness of human tutoring, intelligent tutoring systems, and other tutoring systems.
\newblock \emph{Educational psychologist}, 46\penalty0 (4):\penalty0 197--221, 2011.

\bibitem[Woolf(2010)]{woolf2010building}
B.~P. Woolf.
\newblock \emph{Building intelligent interactive tutors: Student-centered strategies for revolutionizing e-learning}.
\newblock Morgan Kaufmann, 2010.

\bibitem[Yaniv and Kleinberger(2000)]{Yaniv_Kleinberger_2000}
I.~Yaniv and E.~Kleinberger.
\newblock Advice taking in decision making: Egocentric discounting and reputation formation.
\newblock \emph{Organizational behavior and human decision processes}, 83\penalty0 (2):\penalty0 260–281, 2000.

\end{thebibliography}

\end{document}